\documentclass[letterpaper, 10 pt, conference]{ieeeconf}  % Comment this line out if you need a4paper

\IEEEoverridecommandlockouts                              % This command is only needed if
\usepackage{graphicx} % for pdf, bitmapped graphics files

\usepackage{tabularx}

\title{\LARGE \bf
On automatic extraction of on-street parking spaces using park-out events data
}

\author{J.-Emeterio Navarro-B., Martin Gebert and Ralf Bielig % <-this % stops a space
\thanks{Jesús-Emeterio Navarro-Barrientos, Martin Gebert and Ralf Bielig are with MBition GmbH, Daimler AG, Dovestrasse 1, 10587 Berlin, Germany. 
{\tt\small Corresponding author: jesus\_emeterio.navarro-barrientos@daimler.com}}%
}

\usepackage{fancyhdr}

\fancypagestyle{specialfooter}{%
  \fancyhf{}
  
  \fancyfoot[C]{
\small © 2021 IEEE.  Personal use of this material is permitted.  Permission from IEEE must be obtained for all other uses, in any current or future media, including reprinting/republishing this material for advertising or promotional purposes, creating new collective works, for resale or redistribution to servers or lists, or reuse of any copyrighted component of this work in other works.
}
}

\begin{document}

\maketitle
\thispagestyle{specialfooter}
\pagestyle{empty}

%%%%%%%%%%%%%%%%%%%%%%%%%%%%%%%%%%%%%%%%%%%%%%%%%%%%%%%%%%%%%%%%%%%%%%%%%%%%%%%%
\begin{abstract}
This article proposes two different approaches to automatically create a map for valid on-street car parking spaces. 
For this, we use car sharing park-out events data.
The first one uses spatial aggregation and the second a machine learning algorithm. For the former, we chose rasterization and road sectioning; for the latter we chose decision trees. We compare the results of these approaches and discuss their advantages and disadvantages.
Furthermore, we show our results for a neighborhood in the city of Berlin and report a classification accuracy of 91.6\% on the original imbalanced data.
Finally, we discuss further work; from gathering more data over a longer period of time to fitting spatial Gaussian densities to the data and the usage of apps for manual validation and annotation of parking spaces to improve ground truth data.
\end{abstract}

%%%%%%%%%%%%%%%%%%%%%%%%%%%%%%%%%%%%%%%%%%%%%%%%%%%%%%%%%%%%%%%%%%%%%%%%%%%%%%%%
\section{INTRODUCTION}
\label{sec:intro}

\begin{figure}[thpb]
  \centering
  %\framebox{\parbox{3in}{
    \includegraphics[width=1.63in]{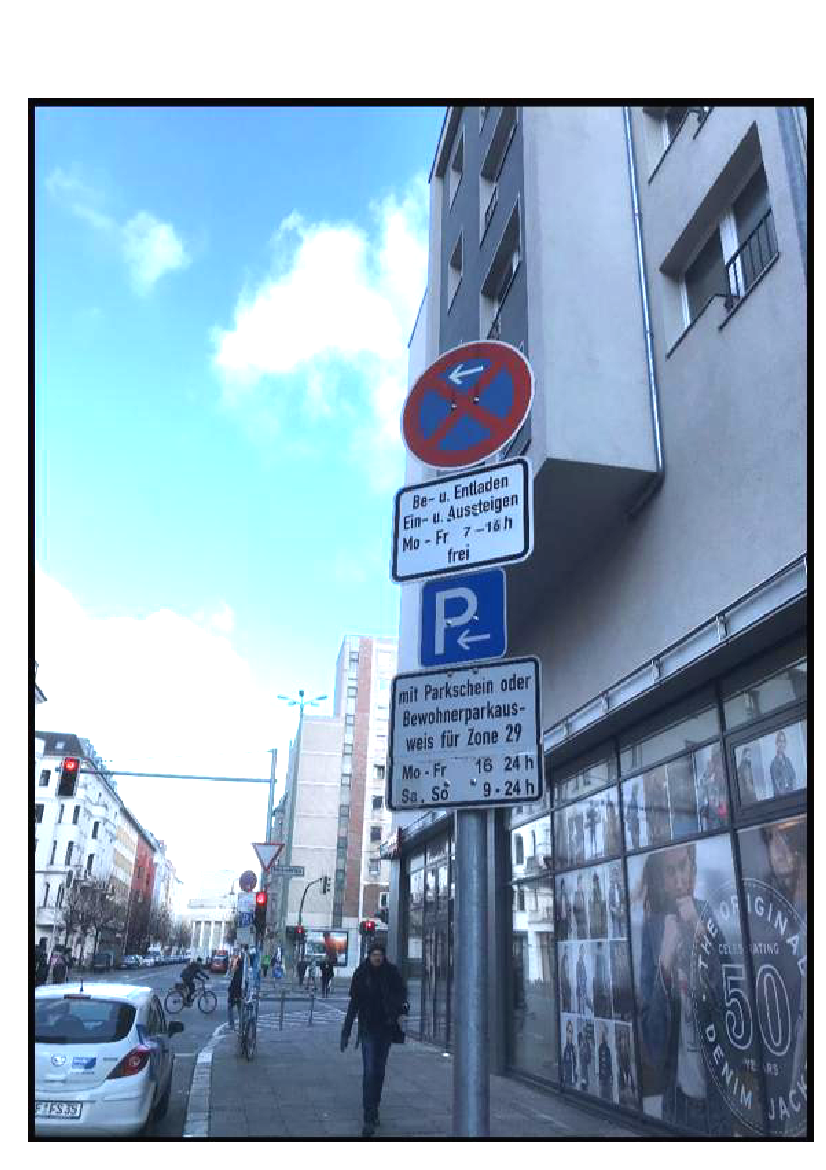}
    \includegraphics[width=1.6in]{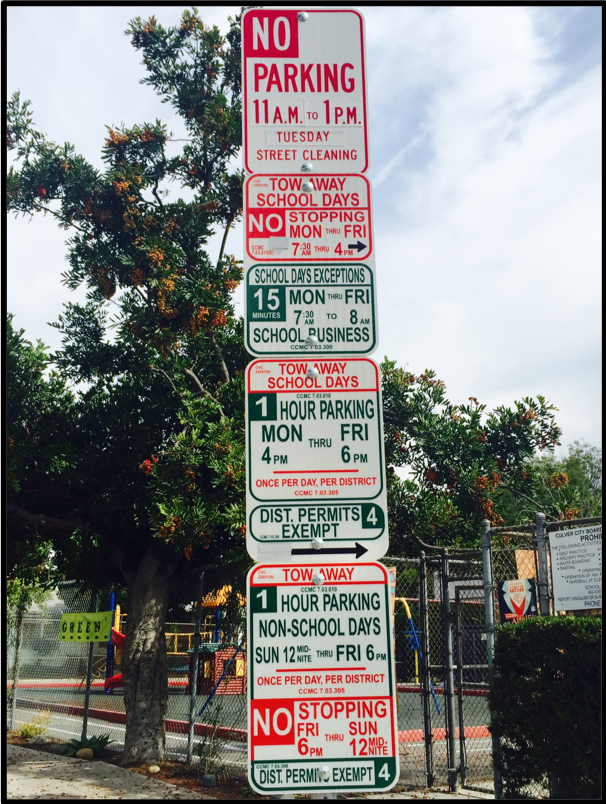}
  %}}
  \caption{Parking signs in Germany (left) and in the US (right).}
  \label{fig:motivationParkingShields}
\end{figure}

One of the biggest difficulties when traveling by car is finding a suitable parking space.
According to a recent parking study \cite{Cookson2017}, car drivers in Germany spend 1.9 billion hours a year looking for a suitable parking space.
In Berlin for example, the average time spent by a driver searching for a parking space is of 62 hours per year, which is equivalent to more than one thousand Euro fuel consumption per year.
This has a considerable impact on the environment, as it also consumes 3.2 billion liters of fuel every year.
Especially in large cities, it is difficult to find parking spaces, as they are usually only available in small quantity and often entail enormous costs.
%On average, a parking space in downtown Berlin costs 4 Euro, in other cities like Stuttgart the price is even higher.
Additionally, penalty costs for wrong parking amounts to 380 million Euro every year.
In this paper, we focus on the latter problem, wrong parking, in other words, on-street valid parking spaces.
Many drivers face the problem of not being sure if they are allowed to park in a given on-street space, independently if they have to pay or not a parking fee.
The problem can be lack of good visibility, lack of time to find and read traffic parking signs or unclear available information about the parking rules for that particular street.
%In some cities, like in Florence Italy, there is a day every month that the street is cleaned-up and no cars should park on the street, if they do, they are towed.
Moreover, some parking signs may be too long or complicated to understand, specially for inexperienced drivers or driving in foreign countries.
%skip motivation Fig. to fit 6 pp.
For example, Fig.~\ref{fig:motivationParkingShields} shows two different shields from Germany and the US, resp.
Many drivers would have some difficulties when trying to understand all these signs.
To counteract the parking problem, governments and urban designers have proposed different solutions.
%skipping fig, to fit to 6 pp
%Fig.~\ref{fig:visualParkingGuide} (left) shows some visual help located on the street by Los Angeles Transportation Department.
%The parking chart-like sign is available also on mobile devices.
%Another example is
For example, the project \emph{MITOS}\footnote{http://mitos.smartsantander.eu} in Spain, has implemented a real-time parking system with different type of sensors, like Radio-Frequency IDentification (RFID) sensors.
Data from {\it MITOS} has been used for different research investigations to investigate the feasibility of predicting parking availability, for example using Neural Networks for Time Series Prediction~\cite{Vlahogianni2015}.
Other approaches consider V2V (Vehicle-2-Vehicle) communication among vehicles to collaborate on finding available spots~\cite{Higuchi2019}. 
Some of these approaches use sensors to measure the surroundings to identify free parking spots by measuring the distance from the vehicle to the nearest roadside obstacle~\cite{Roman2018}. 
Other approaches use video cameras and communication platforms to detect parking spaces with high precision rate~\cite{Tatulea2019}.
Recent research papers show also approaches to localize parking cars using satellite images~\cite{Purahoo2019,Zambanini2020}.
The authors describe a complete processing pipeline for raw satellite images including a layer generated from Open Street Model vector data.
An adapted Faster R-CNN oriented bounding box detector is used for the prediction and tested with a dataset of Barcelona.

Several companies provide already solutions for the parking problem, for example: Easypark, INRIX, Parkopedia, Parknav, Parkbob, Spotangels, etc.
Many of them use digital paid parking, digital law enforcement, automated in-car payments, etc.; to collect different data about parking spaces like location, parking types, restrictions and payment options.
However, to our knowledge, all describe valid parking up to a street level and not to a higher resolution like individual parking spaces.
%Usually, in their business models, they sell these collected and pre-processed data to other companies and government institutions.
Some other companies provide solutions displaying parking space data directly to mobile devices.
% For example, ParKing by Karlsruhe Institute for Technology (KIT) and Parkineers by Robert Bosch GmbH {\it (http://parking-app.de)}
%, see Fig.~\ref{fig:visualParkingGuide} (right). _%skipping fig. to fit 6 pages
% In these projects, the authors developed a gamified crowdsourcing application to share parking space information creating an interactive map of parking spaces.
% This app allows the driver to enter or edit information of on-street parking spaces in a playful way.
% Unfortunately, the applications are not available anymore.
For example, \emph{Polis Assist and Coord}\footnote{http://www.polisassist.com/ and  https://www.coord.com/}, provide similar solutions for some cities in the US. However, the business model of such apps leads to some limitations, e.g. low availability or focus on paid parking spots.
%Note also that some of these previous mentioned applications are apparently not anymore available online, for example Parkineers {\it (https://parkineers.com/)}, showing the difficulty to establish a solution in the market. %skipping to fit 6 pp.
%skipping to fit 6pp.
% \begin{figure}[thpb]
%   \centering
%   \framebox{\parbox{3.2in}{
%   \includegraphics[width=1.4in]{visualParkingGuide}
%   \includegraphics[width=1.8in]{parkineers}
%   }}
%   \caption{Parking guide sign in Los Angeles (left) and Parkineers app.}
%   \label{fig:visualParkingGuide}
% \end{figure}

In the field of automatic mapping using GPS data, several algorithms for map inference have been proposed.
One of these approaches uses a one-to-one bottleneck matching distance where sample points on the generated map can be considered as marbles and the ground truth map as holes~\cite{Ahmed15}.
In their approach, if a marble lands near a hole it falls and is a match.
Counting unmatched and empty holes quantifies the accuracy of the map. 
A similar approach considers rasterization for automatic extraction of road networks using GPS data from vehicles~\cite{Lima2009}.
In this article, the authors used 30 million GPS points to construct the road map of Arganil, Portugal.
This approach is inexpensive and easy to update.
The only disadvantage is its accuracy.
Other approaches involve point clustering based algorithms using for example Kernel Density Estimation with both Voronoi diagrams \cite{Davies2006} or skeleton methods \cite{Biagioni2012,Yu2017}, respectively.
In ~\cite{Gao2019}, the authors used data from parking violation tickets for New York City to train six machine learning algorithms.
%The authors report in their results, that the decision tree ended up with a very complex structure.
The authors report that random forest outperforms all other models with a high accuracy score of 88\%.
The authors also address the issue of imbalanced training, between the positive and negative class for parking legality (validity), leading to the problem of getting excellent accuracy only in the dominating class distribution. 
%% new ->
Moreover, other researchers have tried to aggregate parking availability information from different sources and provide estimates for parking availability based on spatial methods and evaluated using data from \emph{SFpark} project of San Francisco system for managing both on and off-street parking \cite{Bock2016}.
%% <-

The main goal of this article is to investigate the feasibility and accuracy of different methods for the automatic mapping of valid parking spaces.
It is not the goal of this paper to publish a final map, but to propose a new method to produce accurate dynamic parking spaces maps, given enough data is provided.
Thus, this article proposes to fill the gap of automatically determining on-street parking spaces, based on previous {\it Park-Out Events (POEs)}.
We define a {\it(POE)}, as the event whenever an ignition start from a given free-floating fleet operator is detected in a backend server together with the GPS position of the car. 
Free-floating mode is needed for our approach, where cars are freely parked in public spaces within the operational area, for example \emph{SHARE NOW}\footnote{http://www.share-now.com/}.
%<-new
For this, we propose in this paper to use these \textit{POEs} to create digital parking space maps using either spatial aggregation or decision trees.
By this means, we produce a weighted geometry of parking spaces, diluting errors from the GPS receivers in order to provide accurate and automatic construction of valid parking space maps.
To the best of our knowledge, currently no approach exists for extracting valid on-street parking spaces from POEs using either rasterization or decision trees.

\section{DATA UNDERSTANDING}

\begin{figure}[thpb]
   \centering
%   \framebox{\parbox{3.3in}{
   \includegraphics[width=3.3in]{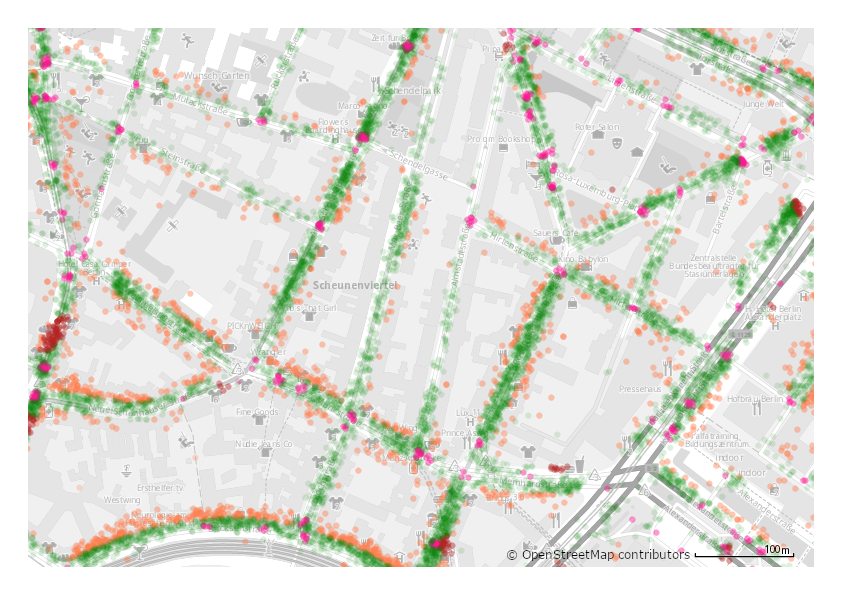}
%   }}
   \caption{Car sharing parking data for a neighborhood in the center of Berlin.}
   \label{fig:poesBerlinMitte}
\end{figure}

We collected car sharing data from car2go (now SHARE NOW) fleet with approximately two million POEs for a period of 6 months, between February and July of 2017, for the city of Berlin.
The data we collected consists in anonymised GPS coordinates of POEs, parking duration, time and date.
Any data that was used for this prototype was pseudo-anonymised without any means available to reconnect a single driver or vehicle with a POE.
Several studies can be found that analyse car sharing data like booking data, travel paths, temporal usage patterns and parking times \cite{Boldrini2016, Schmoeller2015}.
Note that due to copyright and privacy issues, we couldn't publish the results of our investigations until now.
However, note that this data and the results reported in this paper are still up-to-date, given that on-street parking spaces do not change drastically every year.
Fig.~\ref{fig:poesBerlinMitte} shows a snapshot of the data for the  neighborhood of {\it Hackescher Markt} in Berlin with approximately forty thousand POEs.
In this map, we can see that a road network emerges automatically from the raw data, which gives already a visual validation of the data. 
%new
Note that POEs near junctions are colored pink, prohibited parking spots like taxi stands are colored red and all the rest POEs (considered preliminary as valid) are colored green.

\begin{figure}[thpb]
   \centering
%   \framebox{\parbox{3.1in}{
   \includegraphics[width=3.3in]{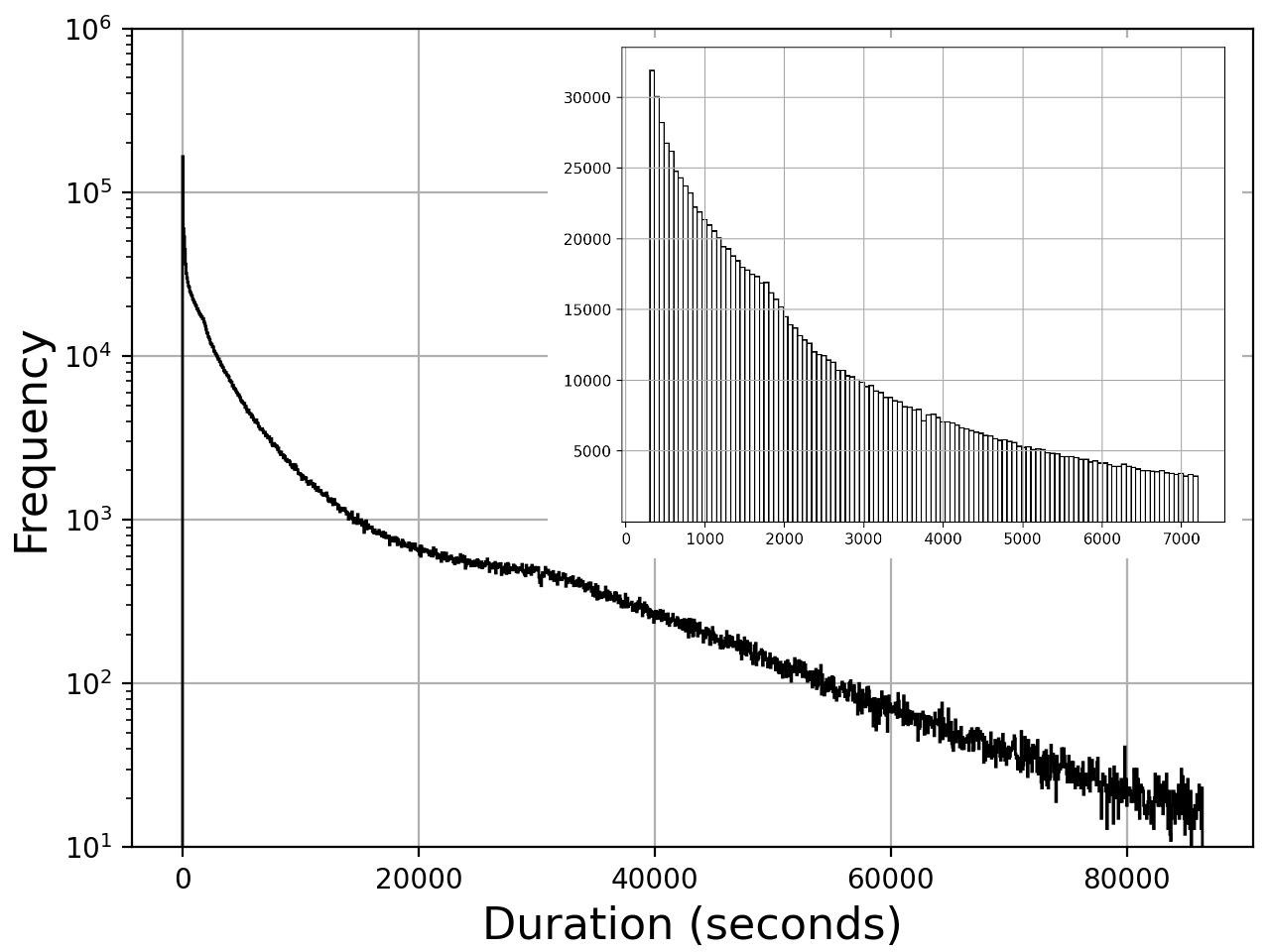}
%   }}
   \caption{Parking duration histogram for car sharing cars for 24 hours. Inset shows a snapshot data from 5 minutes to 2 hours.}
   \label{fig:histosParkingDurationcar-sharing}
\end{figure}

To assess the quality of POEs data, we focused on parking duration and used simple statistics to filter out false positive, for example invalid parking events with duration of less than five minutes.
%new
Fig.~\ref{fig:histosParkingDurationcar-sharing} shows frequencies and parking duration for the car sharing data.
Note that the parking duration follows an inverse gamma and an exponential decay (log-normal) distribution as reported in \cite{Boldrini2016}.
This statistical analysis supports the visual validation shown in Fig.~\ref{fig:poesBerlinMitte}.
The parking duration for the histogram was set to maximal 24 hours and the histogram in the inset shows the interval from 5 minutes to 2 hours.
However, approx. 160.000 POEs have a duration of less than 5 minutes, which denotes a very short parking times to consider them valid parking events.
Furthermore, we used \textit{OpenStreetMap (OSM)} to filter out GPS points located on unfeasible parking zones, like inside parks, schools, private back yards or buildings.

\section{MODELING AND EVALUATION}
We implemented the following two different approaches:
\begin{itemize}

\item {\it Spatial Aggregation}: collect POEs together and map them to spatial statistical structures defining valid parking spaces, providing relatively simple results, easy to interpret and easy to implement.
\item {\it Machine Learning}: collect for each POE, information about road properties relevant to that POE and train a machine learning algorithm to predict if a new POE is a valid parking space or not.

\end{itemize}

\subsection{Spatial Aggregation}
\label{sec:aggregation}

We implemented two types of spatial aggregation: i) uniform grid rasterization and ii) road sectioning.

\subsubsection{Uniform grid rasterization}
\label{sec:rasterization}

\begin{figure}[thpb]
   \centering
   %\framebox{\parbox{3.1in}{
   \includegraphics[width=3.0in]{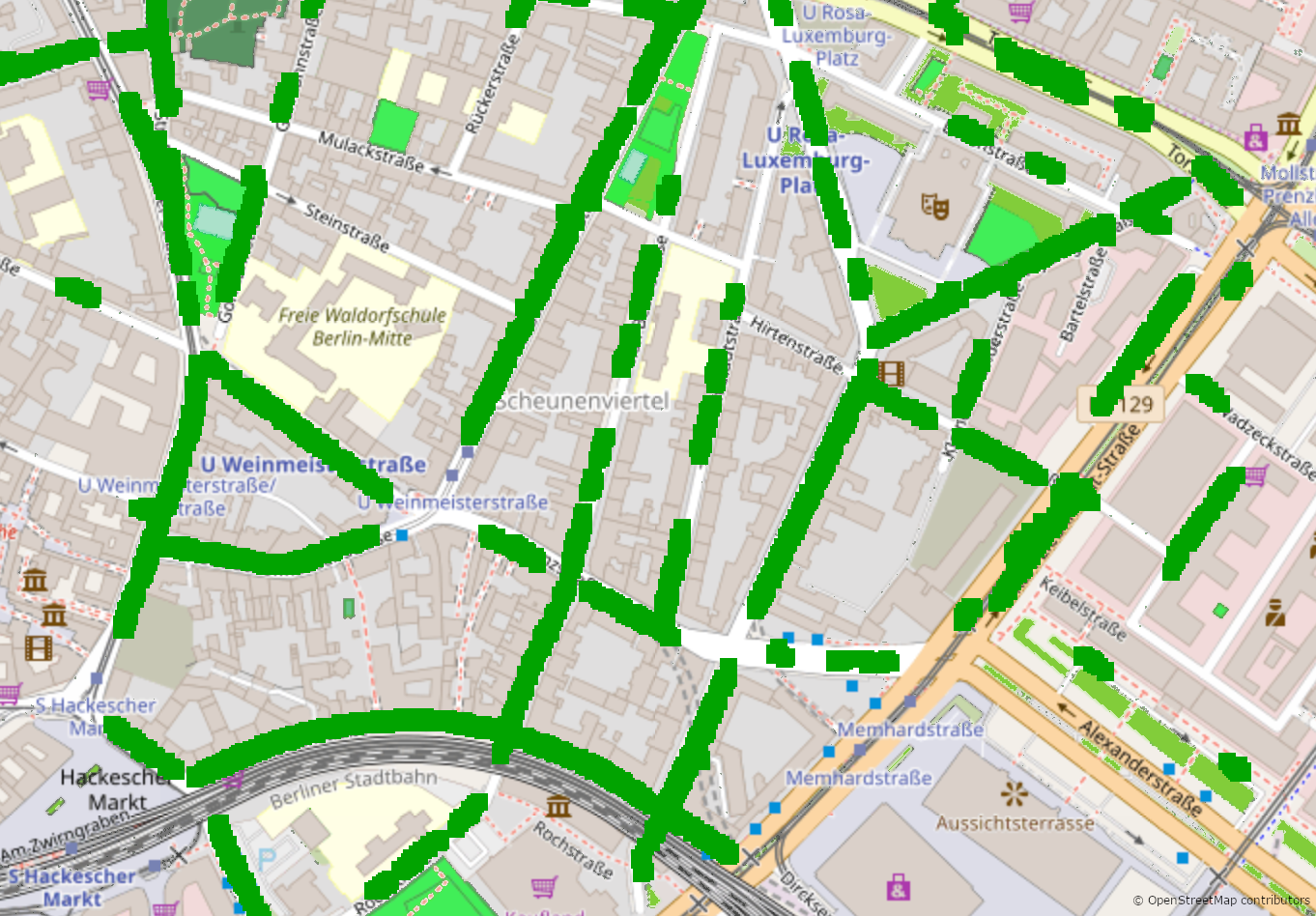}
   %}}
   \caption{Spatial aggregation of PoEs using {\it uniform grid rasterization}.}
   \label{fig:rasterGreenFineTunningResult}
\end{figure}

In a first step, we create a {\it high-resolution raster} layer of the size of the data bounding-box with a fixed cell size of 5-meter width.
We created the grid using {\it OpenStreetMap (OSM)} Slippy Map Tiles, note that we obtain rectangles and not squares, because a degree of latitude does not equal a degree of longitude.
In a second step, we place the set of GPS lat/long coordinate points into the raster and count the number of GPS coordinate points in each cell.
Note that we need to normalize the counter values on each cell with the total number of points for a large number of cells in the neighborhood using a raster with lower resolution.
Concretely, we used a {\it low-resolution raster} with a 500-meter-resolution.
For this, we iterate over the raster with broader resolution and those parking-cells inside a normalization window cell should be normalized to the total number of POEs on that normalization window cell.
By this means, the probability of a valid parking space in a given cell is proportional to the counter value in that cell.
Note that another option would be to iterate over the whole high-resolution raster and normalize each cell according to the number of events on the 1st, 2nd, 3rd, or nth-neighbors.
In other words, to compute a moving normalization window for each cell, however, this process is time-consuming for the size of our raster. Fig.~\ref{fig:rasterGreenFineTunningResult} shows the result of this process again for the neighborhood of {\it Hackescher Markt}.
For this map, we filtered out all cells with low parking events.
This means, we show in green only those cells in the high-resolution raster (5 meters width) with counter values above the average on their low-resolution raster (500 meters width).
This process clearly shows no valid parking for one-way streets where on-street parking is not allowed, like Mulackstraße and Steinstraße located in the center of the map.
Bus stops are also correctly not marked with green boxes.

\begin{figure}[thpb]
   \centering
   %\framebox{\parbox{2.7in}{
   \includegraphics[width=3.0in]{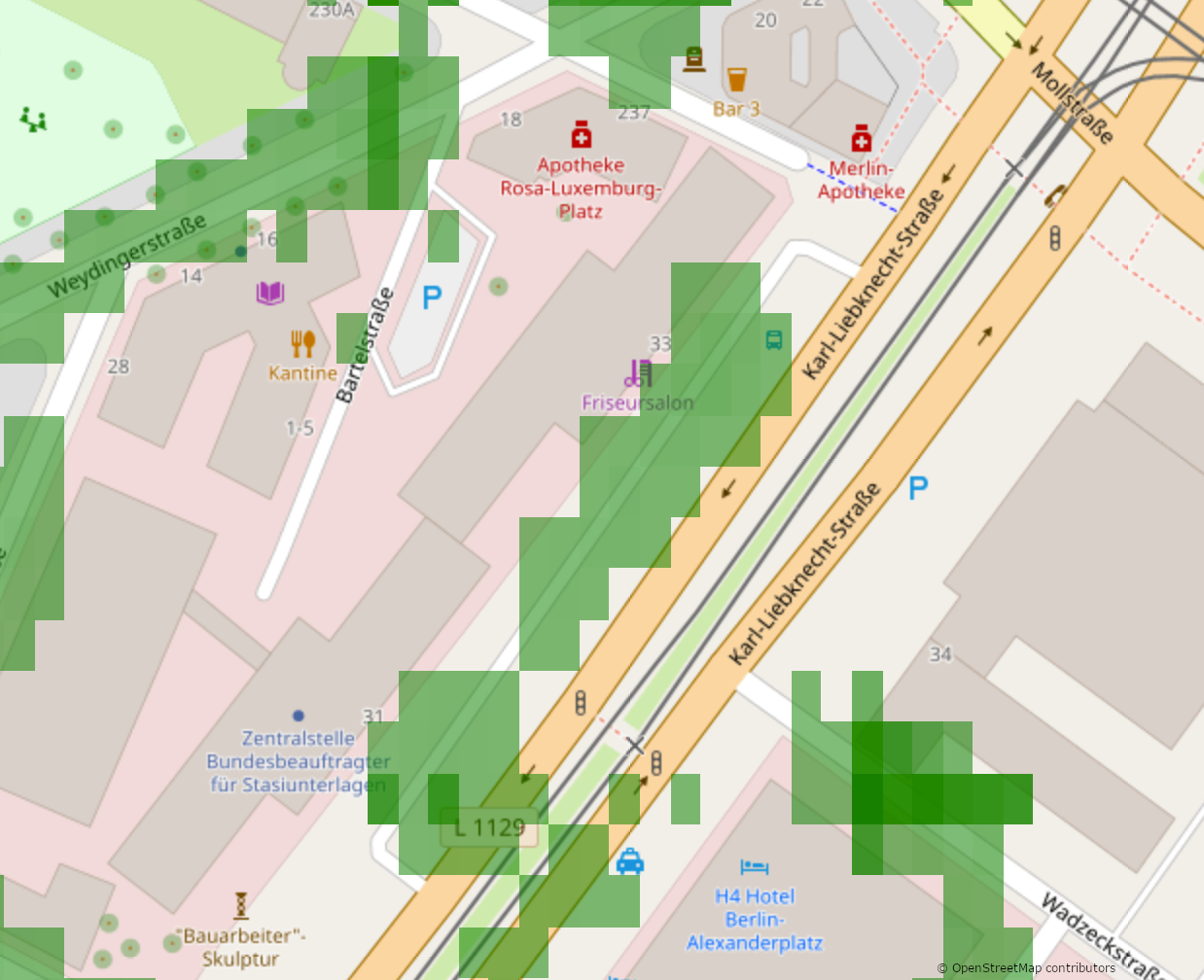}
   %}}
   \caption{Normalized spatial aggregation of PoEs, the darker the color the more probable the validity of the parking space.}
   \label{fig:rasterDarkGreenKarlLieb}
\end{figure}

The normalized counter number can be later mapped to a given color intensity, where the more GPS points in the cell, the darker the cell.
This level of color intensity in a given cell can be then interpreted as a probability of a valid parking space in the cell.
The result allows visual identification of on-street valid parking spaces.
In this manner, we shift the responsibility to the user to determine or decide if a parking space is valid or not based on the visual interpretation of the color intensity of the cell.
Fig.~\ref{fig:rasterDarkGreenKarlLieb} shows the result of this process for a stretch of the street of Karl-Liebknecht-Straße in Berlin-Mitte.
Note that intersections and  entrances to big buildings are correctly filtered out showing only those valid on-street parking spaces with different green intensities, where the darker the color the more probable that the parking space is valid.

\subsubsection{Road sectioning}

\begin{figure}[thpb]
   \centering
   %\framebox{\parbox{3.1in}{
   \includegraphics[width=3.3in]{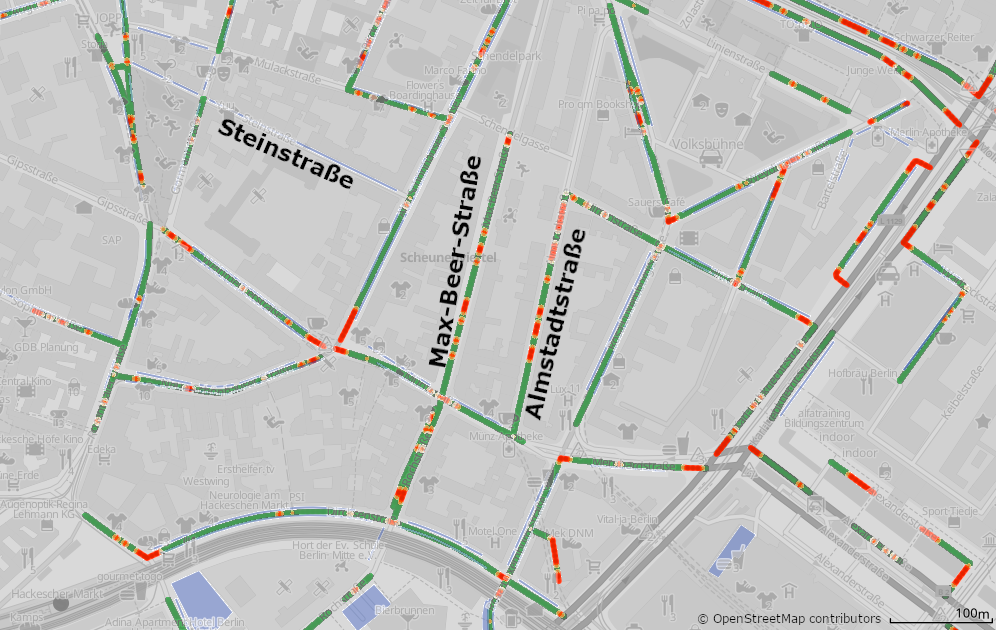}
   %}}
   \caption{Spatial aggregation of PoEs using {\it road sectioning}.}
   \label{fig:car-sharing_restrictions}
\end{figure}

In a first step, we use {\it OSM} to find all car-drivable roads inside the data bounding-box.
In a second step, we divide each road into a variable number of sections, depending on the type of road element.
This approach has been shown to be more useful than dividing a road into sections with fixed constant length; a fixed-length section may have variable geometry within it, like curves~\cite{Koorey2009}.
The road elements and road sections were determined also using data from OSM.
In a third step, we map the GPS lat/long coordinate points of the POEs to their corresponding nearest road section.
%Note that points farther than 10 meters were filtered out as they could be points already from parking areas inside buildings or no on-street parking zones.
In a fourth step, we count the number of events on each road section and normalize the counter values by the length of the road section.
Finally, we split road sections into fixed constant segments of 5 meter size and compute the road segment POEs load ratio, normalizing it by the number of segments on a road section.
For simplicity, we did not differentiate between left and right sides of a road.
Fig.~\ref{fig:car-sharing_restrictions} shows the result of this process again for the neighborhood of {\it Hackescher Markt}.
In green (valid parking), road segments with load ratios above the average load ratio on their road section. 
In red (invalid parking), road segments with no POEs on their road section.
For comparison purposes, straight lines in blue are labels for valid on-street parking from a third-party provider. The company's name is not disclosed due to privacy concerns.
Note that some blue labels from our third party provider are wrong.
For example, on Sophienstraße, the thrid-party provider shows that parking is allowed, while actually this is a narrow one-way street with no valid on-street parking.
Contrary situation is with streets Max-Beer-Straße, Rochstraße or Almstadtstraße.
These streets have no blue labels, indicating no valid on-street parking; while actually on-street parking is allowed on these streets.

%skipping as not visible and to fit 6pp
% For comparison purposes, straight lines in blue are labels for valid on-street parking from a third-party provider. 
% The company's name is not disclosed due to privacy concerns.
% Note that some blue labels from our third party provider are wrong,
% For example, on Sophienstraße, the thrid-party provider shows that parking is allowed, while actually this is a narrow one-way street with no valid on-street parking.
% % HERE!!! maybe skip this sentence to fit 6pp.
% Contrary situation is with streets Max-Beer-Straße, Rochstraße or Almstadtstraße.
% These streets have no blue labels, indicating no valid on-street parking; while actually on-street parking is allowed on these streets.

\subsection{Machine Learning}
\label{sec:dt}

Among the many different Machine Learning methods that exist like Artificial Neural Networks, Decision Trees, Random Forests and Support Vector Machines, we decided to use Decision Trees because they are relatively fast, require minimum parameter tuning, and produce accurate transparent explanatory models easy to interpret.
A {\it Decision Tree}, classifies instances by following a tree-structure of several {\it if-then} conditional statements, which in combination define the final predicted class of a given instance.
In other words, a decision tree does not base its classification on a simultaneous evaluation of the values of all input features of a given instance, but instead follows a sequential approach. 
%skipping to fit to 6pp
Decision Trees are practical as their are also capable of multiclass-classification as well as regression.%~\cite{Runkler2015}.
%(\cite{Runkler2015},~pp.~102–106). %doesn't look good the pp. w this citation format.
%For many of the alternative approaches this is not the case. 
We focused on producing an understandable prediction model because our high-level goal was to explore the problem space and possible solutions, not to build a fine-tuned highly optimized prediction. 
% HERE!: removed, to keep only 1 DT, to save space.
%For comparison purposes, we built two different decision trees, based on different attributes of our POEs.
%The first tree is based on map data such as the locations of junctions, bus stops, etc.; while the second one is %based on the number of POEs that have been recorded in specific areas.

%\subsubsection{Map features based}

To build our decision tree, in a first step, we enriched the data about our POEs by combining it with relevant OSM data. For each POE, we computed the attributes shown in table \ref{tab:tree1}. Apart from $side_type_input$, all are distances to various parking-related areas. 
Next, we split our available POEs randomly into a training dataset, consisting of 80\% of all POEs, and a test dataset with the remaining 20\% POEs.    
\begin{table}[htbp]
\begin{tabularx}{\columnwidth}{ |p{2cm}|X| } 
 \hline
 {\it duration} & parking duration in seconds \\ 
 \hline
 {\it side\_type\_input} & on which side of the road the car was parked \\ 
 \hline
 {\it noparking\_dist\_input} & distance to closest known no-parking area (e.g. bus lanes, gas stations, ...) \\ 
 \hline
 {\it parking\_dist\_input} & distance to closest known parking area (e.g. parking garages) \\ 
 \hline
 {\it road\_dist} & distance to closest road \\ 
 \hline
 {\it freeway\_dist} & distance to closest freeway \\ 
 \hline
 {\it gas\_dist} & distance to closest gas station \\ 
 \hline
 {\it buslane\_dist} & distance to closest bus lane \\ 
 \hline
 {\it junction\_dist} & distance to closest junction \\ 
 \hline
 {\it center\_dist} & distance to center of Berlin \\ 
 \hline
\end{tabularx}
\caption{Computed Attributes for the Map-Based Decision Tree}
\label{tab:tree1}
\end{table}

We selected a small area of Hackescher Markt of manageable size and built a {\it ground truth (GT)} dataset for it.
% HERE!!! TODO: commented for blind-review
%For this, we used an app developed for iOS to extract and tag openly accessible on-street park data from the open source database OSM ~\cite{Grasser2018}.
%Based on this, 
We manually created the GT data for the selected area, which maps every position of this area to either the label {\it noParking} or {\it yesParking}. 
%By this means, we also contributed to OSM with these tags.

\begin{figure*}[thpb]
    \begin{center}
   \includegraphics[width=1.0 \textwidth]{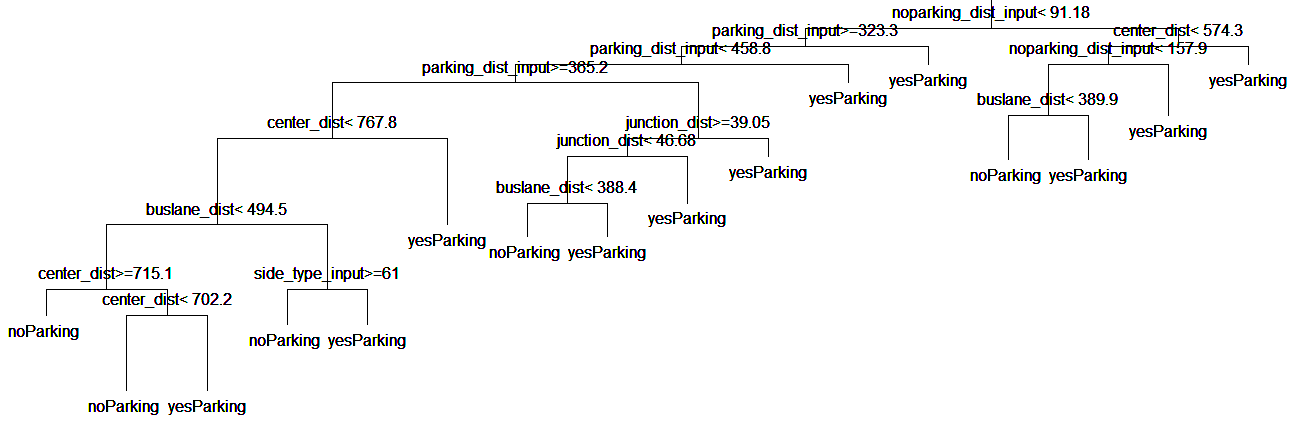}
   \caption{Snapshot of map features based decision tree for predicting POEs.}
   \label{fig:dt}
   \end{center}
\end{figure*}

Using the training and GT datasets, we used standard algorithms to compute a decision tree that predicts the correct parking label.
A snapshot of the resulting decision tree is shown in Fig.~\ref{fig:dt}.
Each inner node of the tree is labeled with a condition. If this condition is true for a specific POE, the algorithm {\it "moves"} to the left side child node; otherwise it moves to the right side child node. Once the algorithm reaches a leaf, the POE is classified according to the label of that leaf.

%skipped to fit 6pp
% One notable detail about the shown decision tree is that POEs which correspond to the left-most leaf are classified as {\it noParking} solely based on their distance to the center of Berlin being between 988.6 and 1063.0 meters. This is obviously not a classification rule that would hold in general, i.e. for other cities or even other parts of Berlin. 
% Instead, this is a clear sign of overfitting. Because our training data covers such a small area, just by chance there happens to be a distance from Berlin center that uniquely matches to one of the {\it noParking} zones in the analyzed area.

%For other parts of the decision tree it is also quite likely that they work only for a very specific area of Berlin and thus are a result of over-fitting. 

We subsequently labeled the POEs in the test data set using our generated decision tree and then compared the predicted labels to the GT. We found that 91.6\% of the predicted labels turned out to be correct. 
Note that due to how our POEs were generated in the first place, they are not evenly distributed across the two categories {\it noParking} and {\it yesParking}. Instead, {\it yesParking} POEs make up 83.7\% of the overall dataset. So a very trivial algorithm predicting always the {\it yesParking} label, would be correct in 83.7\% of the cases. In order to put these numbers into perspective we can look at how much the decision tree reduces uncertainty as compared to the trivial model, which is by 100\% - (100\% - 91.6\%) / (100\% - 83.7\%) = 48.5\%. 

\section{CONCLUSIONS AND FURTHER WORK}

In this paper, we proposed to use spatial aggregation or machine learning to find automatically valid on-street parking spaces.
We compared two different forms of spatial aggregation - rasterization and road sectioning. 
Spatial aggregation approaches are easy to implement and to interpret. 
Among the two approaches we tested, aggregation of POEs using road sectioning is a more promising mechanism, as it introduces more information, like road sections.
This excludes garages, backyards and the like non on-street parking spaces.
Rasterization, on the other hand is much easier to maintain.
For example, whenever any new POE is added to its corresponding cell in the raster; the average for the normalization is recomputed and the intensity of the cell is adjusted accordingly.
A change on the road network, wouldn't affect the {\it rasterization} method, whereas the {\it road sectioning} method would have to recompute a large part of the load ratios and normalization.
Our approach using {\it decision trees} provides accurate predictions and models that are easy to explain and to integrate in other models. 
Note that training any machine learning algorithm can be complicated if there is inconsistency data (bad labeling), inaccurate data (for example from GPS reflected signals) or unbalanced observation data.
Moreover, our decision tree had a prediction accuracy of 91.6\%, and reduced uncertainty by 48.5\%.
%; whereas the density based DT had an accuracy of 92.3\% and reduced uncertainty by 52.8\%.
The decision tree was constructed from ten available POE attributes.
%, while the density based tree was constructed from 183 available attributes. 
The more attributes a prediction model is built from, the more prone it is to over-fitting.
This occurs, because a higher number of attributes results in a higher chance of finding an attribute which (just by chance) has values inside some range for {\it noParking} POEs and values outside that range for {\it yesParking} POEs (or vice-versa). 
%Based on this first exploration of the solution space we conclude that the density based DT is more recommendable to use than the map feature based.
%Additional attribute categories and the combination of map data with density data could further improve prediction quality.
It would be also interesting to increase the training set size which might improve accuracy.
We note that it is important to run future analyses on larger areas to avoid over-fitting. 
An important issue is imbalanced data which leads to bias in our prediction model.
To avoid this, it is possible to equalize sample sizes for each class using over-sampling, under-sampling or bootstrapping strategies, to artificially increase/decrease the overall amount of available data \cite{Kohavi1995}.
%, however, this is left for further work.

We discuss now further work ideas for improving some of the approaches we discussed in this paper as well as new possible approaches.
One natural extension is to increase the categorization of parking spaces.
With a large set of GPS POEs, it should be possible to determine not only if an on-street space is valid or not, but also when (day and time) and for how long (maximal parking duration) in case there are time constraints.
%This would be useful, to determine valid parking during day and night. 
% skip?
%In Berlin for example, many night-bus stop zones can be used during the day as normal on-street parking zones.
%For this, the aggregation and machine learning approaches could be perform separately for each day of the week and different day-times.
% skip?
%In this manner, we could have a particular raster model for Saturdays from 6am to 15pm, and the like.

\begin{figure}[thpb]
   \centering
   \framebox{\parbox{3.4in}{
   \includegraphics[width=1.80in]{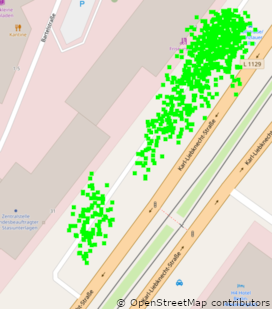}
   \includegraphics[width=1.50in]{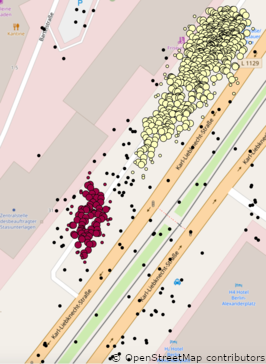}
    \includegraphics[width=3.3in]{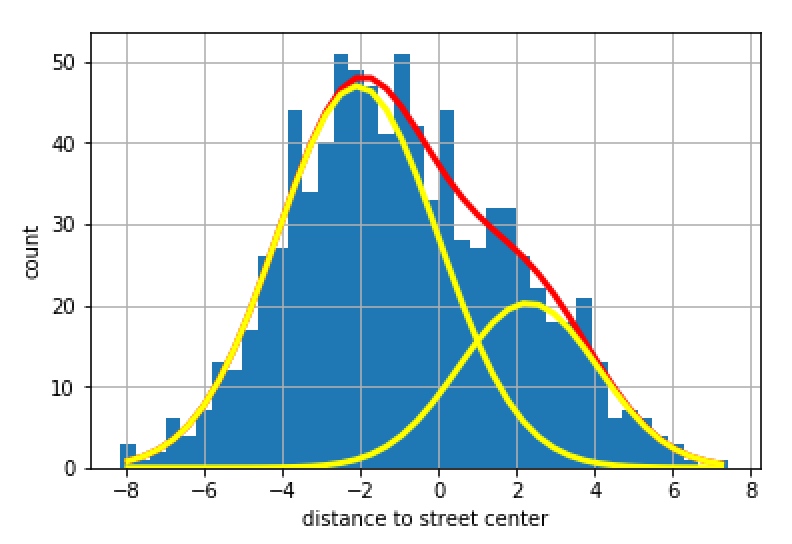}
   }}
   \caption{POEs for a side street in Berlin: raw data (top-left); data fitted to Gaussian density distributions (top-right) and Gaussian density distribution best fitting the data (bottom).}
   \label{fig:densities}
\end{figure}

An interesting and promising approach is to fit spatio-temporal models to parking data \cite{Rajabioun2015}.
We explored the possibility to fit Gaussian density distributions to the POEs.
For this, first we need to fit a line to the POEs of interest.
Secondly, we project the data to the fitted straight line and incrementally find the number of Gaussian distributions that best fit the data. 
By this means, it is possible to find the positions of invalid on-street parking spaces, like entrances, which correspond to the points where the Gaussian distributions intersect.
Fig.~\ref{fig:densities} shows a preliminary result for a side street of Karl-Liebkneckt-Str. in Berlin.
We chose this street as there is a traffic light with pedestrian crossing (no legal parking zone) in the middle of the street.
Fig.~\ref{fig:densities} (top-left) shows the original POEs and (top-right) the result of splitting these by two Gaussian distributions that best fitted the data. 
Fig.~\ref{fig:densities} (bottom) shows these two Gaussian density distributions.
Note that the point where the two distribution intersect match perfectly with the pedestrian crossing located on the street.
Note that this classification using Gaussian density distributions matches also with the result we obtain using {\it uniform grid rasterization} on the same street, shown previously in Fig.\ref{fig:rasterDarkGreenKarlLieb}.
Finally, note that the density distribution approach is not limited to only two density distributions, theoretically, we could match any number of densities up to the number of possible parking spots on a street. 
However, this is out of the scope of this paper.

%removed due to space and also because we already have a similar plot on Karl-Lieb Str.
% Fig.~\ref{fig:maybachufer} shows that ...
% \begin{figure}[thpb]
%   \centering
%   \framebox{\parbox{3.4in}{
%   \includegraphics[width=1.70in]{poesMaybachufer}
%   \includegraphics[width=1.62in]{6densitiesMaybachufer}
%   }}
%   \caption{.}
%   \label{fig:maybachufer}
% \end{figure}

%\subsection{Using special apps for tagging map data}

Further work also includes to improve the ground truth (GT) data.
For this, there are many {\it OpenStreetMap (OSM)} editor-applications available.
The majority of them, however, focus on editing all or just a special part of data from OSM.
Three editors are particularly interesting for this purpose since they provide specific interaction with OSM: the iOS applications {\it Go Map!!}, and {\it PushPin}, as well as the Android application {\it Vespucci}.
Of special interest are the app Parkineers (by Bosch) and ParKing (by KIT). 
These were applications to register and display parking spaces and conditions in a playful way based on a crowd-sourcing approach.
They were available for free in app stores for Android and iOS devices.
However, these applications did not succeed in becoming popular and were discontinued.

Further work also includes to collect more data to improve accuracy. 
Note that we could use Bluetooth connection to get POEs from any car, not necessary from only Mercedes-Benz cars. 
For this, we need to include a background service (with the user's consent) to some already existing mobile application.
Thus, we could collect a POE whenever a user connects to the infotainment system in its car via Bluetooth, all this on the assumption that the driver would be typically starting the car to drive out from a legal parking space.

Finally, we do not have to limit to data from POEs of cars, also previously mentioned data from police parking violations data \cite{Gao2019}, radar sensors from cars or RFID sensors on streets~\cite{Vlahogianni2015} could all together improve the accuracy and predictability of any of the systems we have discussed in this paper.

% skipping fig. to fit 6pp
% Fig.\ref{fig:rasterOnSatellite} shows... using google satellite maps...
% \begin{figure}[thpb]
%   \centering
%   \framebox{\parbox{3in}{
%   \includegraphics[width=3in]{rasterBigBlueSquares}
%   }}
%   \caption{.}
%   \label{fig:rasterOnSatellite}
% \end{figure}

---

\addtolength{\textheight}{-12cm}   % This command serves to balance the column lengths

%%%%%%%%%%%%%%%%%%%%%%%%%%%%%%%%%%%%%%%%%%%%%%%%%%%%%%%%%%%%%%%%%%%%%%%%%%%%%%%%

% new commented for blind-review
\section*{ACKNOWLEDGMENT}

Thanks to our colleagues in MBition GmbH. Frank Bielig for providing car-sharing data, Natalie Grasser for implementing an application for iOS to collect and import ground truth data to OSM and Simon Sala for computing the Gaussian density distributions.

%%%%%%%%%%%%%%%%%%%%%%%%%%%%%%%%%%%%%%%%%%%%%%%%%%%%%%%%%%%%%%%%%%%%%%%%%%%%%%%%

\end{document}